# A scalable approach to modeling on accelerated neuromorphic hardware

Eric Müller [*,†], Elias Arnold [*,†], Oliver Breitwieser [*,†],

Milena Czierlinski [*,†], Arne Emmel [*,†], Jakob Kaiser [*,†],
Christian Mauch [*,†], Sebastian Schmitt [*,‡], Philipp Spilger [*,†],
Raphael Stock [*,†], Yannik Stradmann [*,†], Johannes Weis [*,†],
Andreas Baumbach [†], Sebastian Billaudelle [†], Benjamin Cramer [†],
Falk Ebert [†], Julian Göltz [†], Joscha Ilmberger [†], Vitali Karasenko [†],
Mitja Kleider [†], Aron Leibfried [†], Christian Pehle [†],
Johannes Schemmel [†]

March 21, 2022

Neuromorphic systems open up opportunities to enlarge the explorative space for computational research. However, it is often challenging to unite efficiency and usability. This work presents the software aspects of this endeavor for the BrainScaleS-2 system, a hybrid accelerated neuromorphic hardware architecture based on physical modeling. We introduce key aspects of the BrainScaleS-2 Operating System: experiment workflow, API layering, software design, and platform operation. We present use cases to discuss and derive requirements for the software and showcase the implementation. The focus lies on novel system and software features such as multi-compartmental neurons, fast re-configuration for hardware-in-the-loop training, applications for the embedded processors, the non-spiking operation mode, interactive platform access, and sustainable hardware/software co-development. Finally, we discuss further developments in terms of hardware scale-up, system usability and efficiency.

[*] contributed equally.
[†] Electronic Vision(s), Kirchhoff-Institute for Physics, Heidelberg University, Germany.
[‡] Third Institute of Physics, University of Göttingen, Germany.

# 1 Introduction

The feasibility and scope of neuroscientific research projects is often limited due to long simulation runtimes and therefore long wall-clock runtimes (van Albada et al., 2021). Other areas of neuromorphic research —such as lifelong learning in robotic applications— inherently rely on very long network runtimes to capture physical transformations of their embodiment on the one hand and evolutionary processes on the other. Furthermore, training mechanisms relying on iterative reconfiguration benefit from low execution latencies.

Traditional software-based simulations typically still often rely on general-purpose high-performance computing (HPC) hardware. While some efforts towards GPU-based accelerators provide an intermediate step to improve scalability and runtimes, (Abi Akar et al., 2019; Yavuz et al., 2016), domain-specific accelerators —a subset of them being neuromorphic hardware architectures—, come more and more into the focus of HPC (Dally et al., 2020). Such systems specifically aim to improve on performance and scalability issues — both, in the strong and in the weak scaling cases. Particularly, the possibility to achieve high throughput at low execution latencies can pose a crucial advantage compared to massively parallel simulations.

The BrainScaleS (BSS) neuromorphic architecture is an accelerator for spiking neural networks based on a physical modeling approach. It provides a neuromorphic substrate for neuroscientific modeling as well as neuro-inspired machine learning. Earlier work shows its scalability in wafer-scale applications, emulating up to 200k neurons and 40M synapses (Schmitt et al., 2017; Göltz et al., 2021a; Kungl et al., 2019; Müller et al., 2020b), as well as its energy-efficient application as standalone system with 512 neurons and 128k synapses in use cases related to edge computing (Stradmann et al., 2021; Pehle et al., 2022). Similar to other neuromorphic systems based on the concept of physical modeling, neuroscientific modeling on the BrainScaleS-2 (BSS-2) system requires a translation from a user-defined neural network experiment to a corresponding hardware configuration. Many neuromorphic systems have been providing software solutions to solve this problem and enable higher-level experiment descriptions. We developed a software stack for the wafer-scale BrainScaleS-1 (BSS-1) system covering the translation of user-defined experiments from the PyNN high-level domain-specific description language to a hardware configuration (Müller et al., 2020b). While the BSS-2 neuromorphic architecture hasn't been scaled to full wafer-size yet, other feature additions such as structured and non-linear neuron models and single instruction, multiple data (SIMD) processors already now provide an appealing substrate for modeling of smaller network sizes. In particular, a new challenge is posed by the introduction of SIMD processors in BSS-2 as programmable elements with real-time vectorized access to many observables from the physical modeling substrate. Observables such as correlation sensors are implemented in the synapse circuits, yielding an immense computational power by offloading computational tasks into the analog substrate. Moreover, the configuration space increases significantly: in addition to a static configuration of network topology, the processors allow for flexible handling of dynamic aspects such as structural plasticity, homeostatic behavior and virtual environments enabling robotic or other closed-loop applications, see sections 3.2 and 3.3. This "hybrid" approach requires modeling support in the software stack integrating code generation for the processors as well as mechanisms to parameterize plasticity algorithms and other code parts running on the embedded processors.

We present recent modeling advances on the substrate showcasing new features of the system: complex neurons, structured plasticity, closed-loop sensor-motor interaction, neuro-inspired machine-learning experiments. We demonstrate network-attached accelerator operation as well as standalone operation. We argue that for successful and sustainable advances in the usage of neuromorphic systems a deep integration between hardware and software is crucial on all layers. The complete system —software together with hardware— needs to be explicitly designed to support access with varying abstraction levels: high-level modelers, expert users and component developers possess different perceptions of the system; in order for a modeling substrate to be successful, it has to deliver on all of these aspects.

## 1.1 The BrainScaleS-2 hardware

In this section we introduce the BSS-2 system and highlight the basic hardware design which is guiding the development of the accompanying software stack. For a more in depth description of the hardware aspects of the BSS-2 system refer to Pehle et al. (2022); Schemmel et al. (2020); Aamir et al. (2018).

BrainScaleS is a family of mixed-signal neuromorphic accelerators; analog circuits emulate neuron as well as synapse dynamics in continuous time, while communication of spike events and configuration data is handled in the digital domain. In this paper we focus on the single chip BSS-2 system with 512 neurons and 131 072 synapses circuits, see fig. 1 A. Due to the intrinsic properties of the silicon substrate, the physical emulation of neuron dynamics is 1000 faster than in biological real time. Currently, the BSS-2 ASIC is integrated in a stationary laboratory setup, fig. 1 C, as well as in a portable system, fig. 1 B.

The high configurability of the BSS-2 system facilitates many different applications, section 3. For example, the neuron circuits replicate the dynamics of the adaptive exponential integrate-and-fire (AdEx) neuron model (Brette and Gerstner, 2005) and are individually configurable by a number of analog and digital parameters. By connecting several neuron circuits together to form one logical neuron, more complex multi-compartmental neuron models can be formed and the synaptic fan-in of individual neurons can be increased; a single neuron circuit on its own has access to 256 synapses (fig. 1 D). In addition to the emulation of biological plausible neural networks, BSS-2 also supports non-spiking artificial neural networks (ANNs). This is facilitated by disabling spiking as well as the exponential, the adaptive and the leak current of the AdEx neuron model, turning the neuron circuits into simple integrators. Furthermore, the high configurability allows countering device-specific deviations between analog circuits which result from imperfections during the manufacturing process, see section 2.3.6.

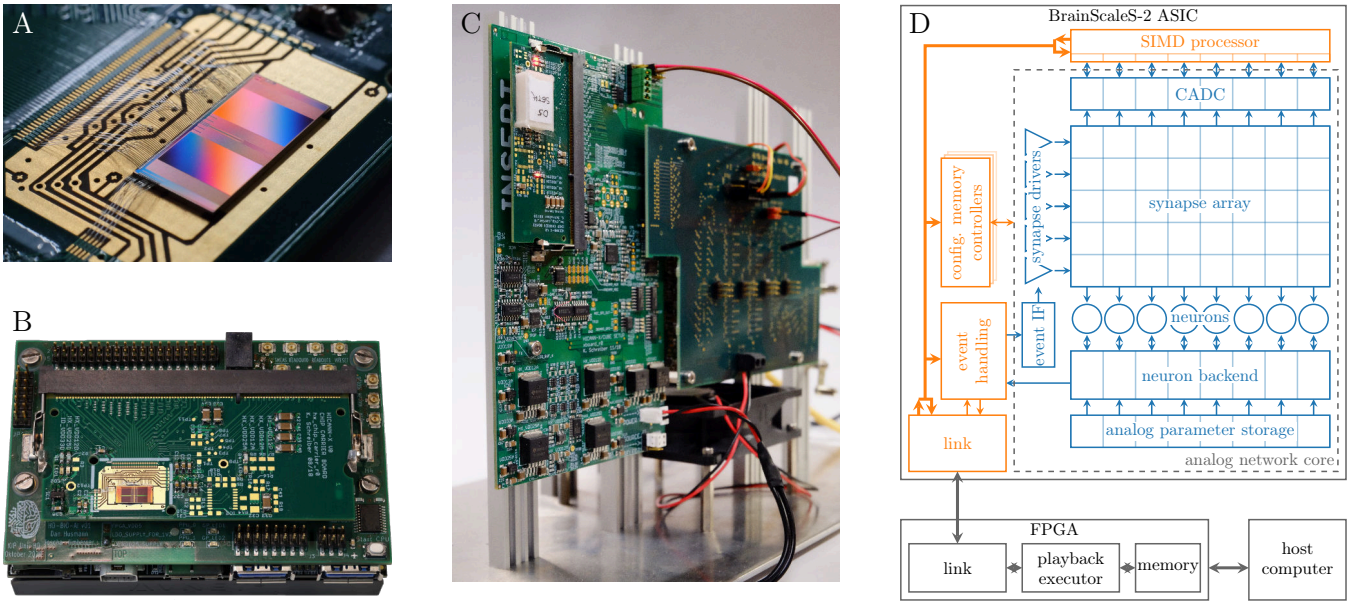

Figure 1: Overview of the BSS-2 system. **(A)** BSS-2 application-specific integrated circuit (ASIC) bonded to a carrier board. The ASIC is organized in two hemispheres each hosting 256 neurons and the accompanying synapse matrix, compare (D). **(B)** Portable BSS-2 system. **(C)** Laboratory setup. **(D)** Overview over the signal flow in the BSS-2 system. The depicted analog neural network core and SIMD processor represent one of the two hemispheres visible in (A), which are mirrored vertically below the neurons.

The digital handling of spike events enables the implementation of various network topologies. All spikes, including external spikes as well as spikes generated in the neuron circuits, are collected in the "event handling" block and subsequently routed off chip for recording or via the synapse drivers and synapses to post-synaptic on-chip partners, compare fig. 1 D. One of the key challenges during experiment setup is the translation of neural networks to valid hardware configurations. This includes assigning specific neuron circuits to the different neurons in the network as well as routing events between neurons, cf. sections 2.3.2 and 2.3.3.

Apart from forwarding spikes, the synapse circuits are also equipped with analog correlation sensors which measure the causal and anti-causal correlation between pre- and post-synaptic spikes. The measured correlation can be accessed by two columnar ADCs (CADCs), which measure correlations row-wise in parallel and can be used in the formulation of plasticity rules, cf. sections 3.2 and 3.3.

An additional analog-to-digital converter (ADC), the so-called membrane ADC (MADC), offers the possibility to record single neurons with a higher temporal and value resolution.

Aside the analog neural network core, two embedded SIMD processors, based on the Power$^{\text{TM}}$ architecture (PowerISA, 2010), which allow for arbitrary calculations and reconfigurations of the BSS-2 ASIC during hardware runtime and are the experiment master in standalone operation. They are equipped with 16 KiB static random-access memory (SRAM) memory each and feature a weakly-coupled vector unit (VU), which can access the hemisphere-local synapse matrix as well as the CADC.

Communication to the BSS-2 ASIC as well as real-time runtime control is handled by a field-programmable gate array (FPGA). It provides memory buffers for data received from a host computer or from the chip, with which it orchestrates experiment executions in real time, see section 2.1. To allow for more complex programs and larger data storage, the on-chip processors can access memory connected to the FPGA.

The software stack covered in this paper handles all the necessary steps to turn high-level experiment descriptions into configuration data, spike stimuli or programs for the on-chip SIMD processor.

In the following we will at first describe the BSS-2 Operating System (BSS-2 OS) in section 2 before showcasing several applications in section 3. We conclude the paper with a discussion in section 4.

## 2 BrainScaleS-2 Operating System

This section introduces key concepts and software components that are essential for the operation of BrainScaleS-2 systems. First, we introduce the workflow of experiments incorporating BSS-2, derive an execution model and specify common modes of operation in section 2.1. Continuing, we give a structural overview of the complete software stack including the foundation developed in (Müller et al., 2020a) in section 2.2. Following this, we motivate key design decisions and show their incorporation into the development of the software stack in section 2.3. Finally, we describe advancements in platform operation towards seamless integration of BSS-2 as an accelerator resource in multi-site compute environments in section 2.4.

### 2.1 Experiment Workflow

Unlike numerical simulations, which are orchestrated as number-crunching on traditional computers, experiments on BSS-2 are more akin to physical experiments in a traditional lab. Just like for these there is a *initialization* phase, which ensures the correct configuration of the system for this particular experiment and a *real-time* section, where the network dynamics are recorded and the actual emulation happens. If multiple emulations share (parts of) the configuration, those experiments can be composited by concatenating the trigger commands for both input and recording (see fig. 2).

The fundamental physical nature of the emulation on BSS-2 requires these control commands to be issued with very high temporal precision as the dynamics of the on-chip circuitry can neither be interrupted nor exactly repeated. To achieve this, the accompanying FPGA is used to play-back a sequence of instructions with clock-precise timing, in the order of 10 ns. In order to limit the FPGA firmware complexity, the play-back unit is restricted to sequential execution, which includes blocking instructions (used for times without explicit interaction), but excludes branching instructions. Concurrently to the FPGA-based instruction sequence execution, the embedded single instruction, multiple data central processing units (SIMD CPUs) can be configured to perform readout of observables and arbitrary alterations to the hardware configuration. This means that conditional decisions, e.g. the issuance of rewards, can be performed either via the SIMD CPU if they are not computationally too complex or via synchronisation with the executing host computer which in the current setup has no guaranteed timing.

The initialization phase typically includes time-consuming write operations to provide an initial state of the complete hardware configuration. This is due to both, the amount of data to be transmitted, e.g. for the synapse matrix, and required settling-time for the analog parameters. Since this can take macroscopic amounts of time, at least around 100 µs due to round-trip latency, around 100 ms for a complete reconfiguration, back-to-back concatenation of real-time executions is needed to keep their timeshare high and therefor the configuration overhead low.

Due to the hardware's analog speed-up factor compared to typical biological processes, a single real-time section can be short compared to the initialization phase. Therefore, we concatenate multiple real-time sections after a single initialization phase to increase the real-time executions' timeshare. In the following, this composition is called execution instance and is depicted in fig. 2.

Alternatively, instead of this asynchronous high-throughput operation, the low minimal latency allows for fast iterative workflows with partial reconfiguration, e.g., iterative reconfiguration of a small set of synaptic weights.

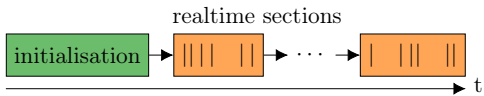

Figure 2: Time evolution of a single execution instance. The initialization is followed by possibly multiple real-time executions with input spike-trains represented by vertical lines.

Based on this we differentiate between three modes of operation. First, in batch-like operation one or multiple execution instances are predefined and run on hardware. Second, in the so-called hardware in-the-loop case hardware runs are executed iteratively where the results of previous runs determine the parameters of successive runs. Last, in closed-loop operation is characterized by tightly coupling the network dynamics of the analog substrate to the experiment controller, either the SIMD CPU or the control host.

## 2.2 Software Stack Overview

Structuring software into well-defined layers is vital for keeping it maintainable and extendable. The layers are introduced and implemented via a bottom-up approach matching the order of requirements in the current stage of the hardware development and commissioning process. This means, that first raw data exchange and transport from and to the hardware via the communication layer is established. Subsequently, the hardware abstraction layer implements translation of typed configuration, e.g. enabling a neuron's event output, to and from this raw data. On this level, the calibration layer allows to programmatically configure the analog hardware to a desired working point. Then, hardware-intrinsic relations between configurables and their interplay in experiments, cf. section 2.1, is encapsulated in a graph structure. Lastly, automated generation of hardware configuration from an abstract network specification enables embedding into modelling frameworks for high-level usage. Figure 3 gives a graphical overview of this software architecture[1].

### 2.2.1 Communication

From the software point of view, the first step to utilize hardware systems is the ability to exchange data. With proper abstraction the underlying transport protocol and technology are interchangeable. Communication is therefore structured into a common *connection* interface *hxcomm*[2] that supports various back-ends.

For most hardware setups, we use a custom, reliable regarding data integrity, transport protocol on top of the user datagram protocol (UDP), *Host-ARQ* provided by *sctrltp*[3]. Additionally, we support connection to hardware design simulations via *flange*[4], compare section 3.6 for both the use during debugging of current and unit testing of future chip generations. Multi-site workflows are transparently enabled already at this level via the micro scheduler *quiggeldy*[5].

### 2.2.2 Hardware Abstraction

A major aspect of any system configuration software is *hardware abstraction*, which encapsulates knowledge about the raw bit configuration, e.g. that bit $i$ at address $j$ corresponds to enabling neuron $k$'s event output. It therefore decouples hardware usage and detailed knowledge about its memory layout, which is an important step towards providing hardware access beyond the group of developers of the hardware. Responsibility of this layer can be compared to device drivers. The layers provide an abstract software representation of various hardware components, such as synaptic weights on the chip or values of supply voltages on the periphery board, as well as their control flow.

---

[1] All the repositories mentioned in the following are available at https://github.com/electronicvisions under the *GNU Lesser General Public License v2/v3*.
[2] *hxcomm* is available at https://github.com/electronicvisions/hxcomm
[3] *sctrltp* is available at https://github.com/electronicvisions/sctrltp
[4] *flange* is available at https://github.com/electronicvisions/flange
[5] *quiggeldy* is available at https://github.com/electronicvisions/hxcomm

Within this category the lowest layer is *fisch*[6] (FPGA Instruction Set arCHitecture), the abstraction of FPGA instructions. Combined with communication software this is already sufficient to provide an interface for prototyping in early stages of system development, i.e., the possibility to manually read and write words at memory locations. With knowledge of the hardware's memory layout this allows specifying addresses and word values directly, e.g. bit $i$ (and all other bits in this word with possibly not related effects) at address $j$ which then enables the neuron $k$'s event output.

The heterogeneous set of entities on the hardware as well as their memory layout is arranged via geometric pattern and contain symmetries, e.g. a row of neurons or a matrix of synapses. An intuitive structure of this fragmented address space is provided by the *coordinate* layer *halco*[7]. It represents hardware components by custom ranged types that can be converted to other corresponding coordinate types, e.g. a `SynapseOnSynapseRow` as a ranged integer $i \in [0, 256)$, that allows conversion to a neuron column, see (Müller et al., 2020a).

A software representation of the configuration space of hardware components is implemented by the *container* layer *haldls*[8]. For example a `NeuronConfig` contains a boolean parameter for enabling the spike output. These configuration containers are translatable (e.g. a neuron container represents one, but not a specific one, of the neurons) and also define methods for de- and encoding between their abstract representation and the on-hardware data format given a location via a supplied *coordinate*. A logical function- instead of a hardware subsystem-centered container collection is implemented by the *lola*[9] layer. For example the `AtomicNeuron` collects the analog and digital configuration of a single neuron circuit, which is scattered over two digital configurations and a set of elements in the analog parameter array.

The *runtime control* layer *stadls*[10] provides an interface to describe timed sequences of read and write instructions of pairs of coordinates and containers, e.g. changing the synaptic weight of synapse $i, j$ at time $t$, as well as event-like response data, e.g. spikes or ADC samples. These timed sequences, also called playback programs, can then be loaded to and executed on the FPGA which records the response data. Afterwards, the recorded data is transferred-back to the host computer.

We track the constitution of all hardware setups in a database, *hwdb*[11]. It is used for compatibility checks between hardware and software as well as for the automated selection of stored calibration data. We also use it to provide the resource scheduling service with information about all available hardware

---

[6] *fisch* is available at https://github.com/electronicvisions/fisch
[7] *halco* is available at https://github.com/electronicvisions/halco
[8] *haldls* is available at https://github.com/electronicvisions/haldls
[9] *lola* is available at https://github.com/electronicvisions/haldls
[10] *stadls* is available at https://github.com/electronicvisions/haldls
[11] *hwdb* is available at https://github.com/electronicvisions/hwdb

systems.

This set of layers is feature-complete to formulate arbitrary hardware-compatible experiments and was used as basis for experiments in Göltz et al. (2021b); Czischek et al. (2022); Klassert et al. (2021); Schemmel et al. (2020); Cramer et al. (2022).

### 2.2.3 Embedded runtime

In addition to the controlling host system, the two SIMD CPUs on the BSS-2 ASIC require integration into the BSS-2 OS. To enable users to efficiently formulate their programs, we provide a development environment based on `C++`. It specifically consists of a cross-compilation toolchain based on `gcc` (GNU Project, 2018) that has been adapted to the custom SIMD extensions of the integrated microprocessors (Müller et al., 2020a). More abstract functionality is encapsulated in the support library *libnux*[12], which provides various auxiliary functionality for experiment design. Moreover, the hardware abstraction layer of the BSS-2 OS (cf. section 2.2.2) supports the SIMD CPUs as an additional cross-compiled target for configuration containers as well as coordinates.

### 2.2.4 Calibration

In order to tune all the analog hardware parameters to the requirements given by an experiment, we provide a calibration framework, *calix*[13]. For example an experiment might require a certain set of synaptic time constants for which analog parameters are to be configured while counteracting circuit inequalities. In section 2.3.6, this layer's design is explained in detail. The `Python` module supplies a multitude of algorithms and calibrations for each relevant component of the circuitry: A calibration provides a small experiment based on the hardware abstraction layer, see section 2.2.2, which is executed on the chip for characterization. An iterative algorithm then decides how configuration parameters should be changed in order to match the measured data with given expectations.

The user-interfacing part provides functions that take a set of target parameters and return a serializable calibration result that can be injected to experiment toplevels, cf. section 2.2.6. Additionally, we have the option to calibrate the analog circuits locally on chip, using the embedded processors. Aside of enabling arbitrary user-defined calibrations, we provide default calibrations for spiking operation, cf. for example sections 3.1 and 3.2, and non-spiking matrix-vector multiplication, cf. section 3.4 for convenient entry. They are generated nightly via continuous deployment (CD).

---

[12]*libnux* is available at `https://github.com/electronicvisions/libnux`
[13]*calix* is available at `https://github.com/electronicvisions/calix`

### 2.2.5 Experiment Description

With rising experiment and network topology complexity, a coherent description ensuring topology and data-flow correctness becomes beneficial. Therefore, a signal-flow graph is defined representing the hardware configuration and experiment flow. Compilation and subsequent execution via the hardware abstraction layer, cf. section 2.2.2, of this graph in conjunction with supplied data, e.g. spike events, then forms an experiment execution. The applied execution model follows the experiment workflow described in section 2.1.

While this aids in construction of complex experiments, detailed knowledge of configuration and its interplay is still required. Solving this, a high-level abstract representation of neural network topology building on top of the signal-flow graph description is developed. An automated translation from this high-level abstraction to a valid hardware configuration is handled by a place-and-route algorithm. This enables hardware usage without detailed knowledge of event routing capabilities and interplay of configuration.

This layer is contained in *grenade*[14], short for GRaph-based Experiment Notation And Data-flow Execution. Its design is explained in detail in section 2.3.2.

### 2.2.6 Modeling Wrapper

Various back-end-agnostic modeling languages emerged to provide access to various simulators or neuromorphic hardware systems to a wide range of researchers. The BSS-2 software stack comprises wrappers to two of such modeling frameworks: PyNN (Davison et al., 2009) via *pyNN.brainscales2*[15] and PyTorch (Paszke et al., 2019) via *hxtorch*[16] (Spilger et al., 2020). Their goal is to provide a common user interface and to embed different back-ends into an existing software ecosystem. This allows users to benefit from a consistent and prevalent interface and integration into their established work-flow. The design of these layers' integration with BSS-2 is explained in detail in section 2.3.4 for PyNN and in section 2.3.5 for PyTorch.

## 2.3 Software Design

We base the full-stack software design on the principles laid out in Müller et al. (2020a). We use `C++` as the core language to ensure high performance and make use of its compile-time expression evaluation and template metaprogramming capabilities. Due to the heterogeneous hardware architecture we employ type safety for logical correctness and compile-time error detection. Serialization support of configuration and control flow enables multi-site workflows as well as archiving of experiments.

---

[14] *grenade* is available at https://github.com/electronicvisions/grenade
[15] *pyNN.brainscales2* is available at https://github.com/electronicvisions/pynn-brainscales
[16] *hxtorch* is available at https://github.com/electronicvisions/hxtorch

In the following, we show enhancements of the hardware abstraction layer, see section 2.2.2, introduced in Müller et al. (2020a) as well as design decisions for the full software stack with high-level user interfaces. First, support for multiple hardware revisions is shown in section 2.3.1. Then, the signal-flow graph-based experiment notation is derived in section 2.3.2. Following, an abstract network description explained in section 2.3.3 closes the gap to the modelling wrappers in PyNN, cf. section 2.3.4 and PyTorch, cf. section 2.3.5. Closing, the calibration framework is described in section 2.3.6.

### 2.3.1 Multi-revision hardware support

As platform development progresses, new hardware revisions require software support. This holds true for both, the ASIC and the surrounding support hardware like the FPGA and system printed circuit boards (PCBs). Additionally, the platform constitution evolves, e.g. by introduction of a mobile system with still one chip but different support hardware or a multi-chip setup.

After a potential development of a second revision, a heterogeneous set of hardware setups may co-exist. For one generation of chips, it is typically possible to combine different revisions with different surrounding hardware configurations, leading to a number of combinations given by the Cartesian product $N = M_\mathrm{ASIC} \times M_{\mathrm{Platform}_1} \times \cdots \times M_{\mathrm{Platform}_P}$, where $M_{\mathrm{Platform}_i}$ is the number of configurations for a given part of the platform, e.g. the FPGA and $M_\mathrm{ASIC}$ is the revision of the BSS-2 ASIC.

We provide simultaneous software support by dependency separation and extraction of common code for each affected component across all affected software layers. This way, code duplication is minimized, maintainability of common features is ensured and divergence of software support is prevented. Moreover, phasing-out or retiring hardware revisions is possible without effecting the software infrastructure of other revisions. The to be implemented software reduces to $N' = M_\mathrm{ASIC} + M_{\mathrm{Platform}_1} + \cdots + M_{\mathrm{Platform}_P}$ constituents, the combinations are rolled-out automatically. We use `C++` namespaces for separation and `C++` templates for common code, which depends on the individual platform's constituents.

### 2.3.2 Signal-flow graph-based experiment notation

As stated in section 2.2.2, the hardware abstraction developed in Müller et al. (2020a) is already feature-complete to formulate arbitrary hardware-compatible experiments. However, it lacks a representation of intrinsic relations between different configurable entities. For example, the hard-wired connections between synapse drivers and synapse rows are not represented in their respective configuration but only given implicitly.

Neural networks are predominantly described as graphs. For spiking neural networks single neurons or collections thereof and their connectivity form a graph (Davison et al., 2009; Gewaltig and Diesmann, 2007; Goddard et al., 2001). In machine-learning, the two major frameworks PyTorch (Paszke et al., 2019) and Tensorflow (Abadi et al., 2015) use a graph-based representation of tensor computation or are moving into this direction (PyTorch's JIT intermediate representation (Facebook, Inc., 2021a) and XLA back end (Facebook, Inc., 2021b; Suhan et al., 2021)).

Inspired by this, we implement a signal-flow graph-based experiment abstraction. A signal-flow graph (Mason, 1953) is a directed graph, where vertices receive signals from their in-neighborhood, perform some operation, and transmit an output signal to their out-neighborhood. We integrate this representation at the lowest possible level to fully incorporate all hardware features without premature abstraction.

For BSS-2, the graph-based abstraction is applied at two granularities, see fig. 4. First, the initial static network configuration as well as virtualised computation using the on-chip embedded processors is abstracted as a signal-flow graph. Second, data-flow between multiple individual real-time experiments distributed over chips and time are described as a graph.

The signal-flow graph representation yields multiple advantages. Type safety in the graph constituents facilitates experiment correctness regarding on-chip connectivity and helps to avoid inherently dysfunctional experiments already during specification. Debugging benefits from visualisation of the graph representation, which directly contains implicit on-chip connectivity. Finally, the signal-flow graph is the ideal source of relationship information for on-chip entity allocation optimization or merging of digital operations.

However, the actual signals are not part of the signal-flow graph representation. They are either provided separately (e.g. external events serving as input), will only be present locally upon execution (e.g. synaptic current pulses) or will be generated by execution (e.g. recorded external events). We implement the experiment workflow described in section 2.1 consisting of an initial static configuration followed by a collection (batch) of time evolutions, see fig. 2.

The signal-flow graph is a recipe for compilation towards the lower-level hardware abstraction layer, cf. Müller et al. (2020a), and eventual execution. The specific implementation of the compilation and execution process is separate from the graph representation in order to allow extensibility and multiple solutions for different requirement profiles. Here, we present a just-in-time (JIT) execution implementation. It supports both, spiking and non-spiking experiments. For every execution instance, the local subgraph is compiled into a sequence of instructions, executed and its results processed in order for them to serve as inputs for the out-neighborhood. While it is feature-complete for the graph representation, it introduces close coupling between the execution on the neuromorphic hardware and the controlling host computer. Host-based compilation can be performed concurrently to hardware execution, increasing parallelism. Figure 5 shows concurrent execution of multiple execution instances (A) and the compilation and execution of a single execution instance (B).

### 2.3.3 Abstract network description

The signal-flow graph-based notation from section 2.3.2 eases creation of correct experiments while minimizing implicit knowledge. However, knowledge of hardware routing capabilities is still required to create a graph-based representation of the hardware configuration which performs as expected. This should not be required to formulate high-level experiments. To close this gap, an abstract representation similar to PyNN (Davison et al., 2009), consisting of populations as collections of neurons and projections as collections of synapses, is developed. Given this description, an algorithm finds an event routing configuration to fulfill the abstract requirements and generates a concrete hardware configuration. This step is called routing. Figure 6 visualizes an abstract network description and one corresponding hardware configuration.

### 2.3.4 Integration of PyNN

When it comes to modeling spiking neural networks, a widely used API is PyNN (Davison et al., 2009). It is supported by various neural simulators like NEST (Gewaltig and Diesmann, 2007), NEURON (Hines and Carnevale, 2003) and Brian (Stimberg et al., 2019), as well as by neuromorphic hardware platforms like SpiNNaker (Rhodes et al., 2018) or the predecessor hardware of BSS-2: BSS-1 (Müller et al., 2020b) and Spikey (Brüderle et al., 2009). With the aim of easy access to BSS-2, we expose its hardware configuration via the PyNN interface. The module `pyNN.brainscales2` implements the PyNN-API for BSS-2. It offers a custom cell type, `HXNeuron`, which corresponds to a physical neuron circuit on the hardware and replicates the `lola.AtomicNeuron` from the hardware abstraction layer, see section 2.2.2. This allows to set parameters directly in the hardware domain and gives expert users the possibility to precisely control the hardware configuration while at the same time take advantage of high-level features such as neuron populations and projections. Figure 7 illustrates how these parameters are available in the corresponding interfaces. An additional neuron type supporting the translation from neuron model parameters in SI units is currently in the planning. Otherwise, the PyNN program looks the same as for any other back end. Since the PyNN-API is free from hardware placement specifications, they are algorithmically determined by mapping and routing in *grenade*, cf. section 2.3.3. This step is performed automatically upon invocation of `pynn.run()`, so that the user is not required to have any particular knowledge about event routing on the hardware. Nevertheless, the interface allows that an experimenter can adjust any low-level configuration aside from neuron parameters and synaptic weights.

To exploit the full potential of the accelerated hardware the software implementation's overhead shall be minimal. Figure 8 presents runtime and memory consumption analysis of the whole PyNN-based stack for a high spike count benchmark experiment. 12 neurons are excited by a regular spike train with 1 MHz frequency and their activity is recorded for one second. These settings are chosen as they roughly equate to the maximum recording rate without loss.

The initial overhead of importing `Python` libraries and setting up the PyNN environment only needs to be performed once for every experiment and is independent of the network topology itself. Run time on hardware is about 1.5 s of which roughly 125 ms are initial configuration and 278 ms are transmission of the input spike train. Post-processing the $1.2 \times 10^7$ received spikes (*fisch* and *grenade*) takes about 1.9 s, i.e., in the same order of magnitude as the actual hardware run. Peak memory consumption is reached during post-processing of results obtained after the hardware execution which corresponds to roughly 3 times the minimum memory footprint of the recorded spike train. With this the stack is well suited to also handle experiments with high spike count without introducing a bottleneck.

### 2.3.5 Integration into PyTorch

To enable access to BSS-2 for machine learning applications, we develop a thin wrapper layer to the PyTorch-API. This extension is called *hxtorch* and was introduced in Spilger et al. (2020) for non-spiking hardware operation emulating analog multiply-accumulate operations and compositions thereof. There, we build on top of the same signal-flow graph experiment description as for the spiking mode of operation, cf. section 2.3.2. Operations are mapped to the hardware size by using temporal serialization and physical concurrency. The PyTorch extension enhances this by automatic gradient calculation for training. Same as PyTorch, we implement a functional API in `C++` wrapped to `Python` (e.g. `hxtorch.matmul` comparable to `torch.matmul`) and add modules/layers on top in `Python` (e.g. `hxtorch.nn.Linear` comparable to `torch.nn.Linear`). In contrast, our operations are quantized to the hardware-intrinsic digital resolution (5 bit unsigned activations, 6 bit weights plus sign bit and 8 bit signed results). Execution on the hardware is performed individually for each operation using the JIT execution, see section 2.3.2.

### 2.3.6 Calibration framework

On BSS-2, there are a multitude of voltages and currents controlling analog circuit behavior. While some of them can be set to default values, most of them require calibration in order to match experiment-specific target values and to counteract device-specific mismatch. Fundamentally, the calibration can be executed on a host computer or locally on chip, using the embedded processors. We provide the `Python` module *calix* to handle all aspects of the calibration process.

Model parameters are calibrated by iteratively adjusting relevant parts of the hardware configuration. As an example, the membrane time constant is controlled by a bias current: In order to calibrate the membrane time constant of all neurons, the neurons' membrane potentials are recorded while they decay back to their resting potential after an initial perturbation from the resting state. We can perform an exponential fit to the recorded voltage trace to determine the time constant and iteratively tweak the bias current to reach the desired target.

The calibration routine of each parameter is encapsulated using an object-oriented API providing a common interface. Mainly, two methods allow the iterative parameter search: one applies a parameter configuration to the hardware, while the other evaluates an observable to determine circuit behavior. An algorithm calculates parameter updates during the iterative search. In each step, the measurement from the calibration class is compared to the target value and the parameter set is modified accordingly.

A functional API is provided for commonly used sets of calibrations, for example for calibration of a spiking leaky-integrate and fire (LIF) neuron. Technical parameters and multidimensional dependencies are handled automatically as required in this case. This yields a simple interface for experimenters for tweaking high-level parameters, while calibration routines for individual parameters remain accessible for expert users.

The higher-level calibration functions save their results in a typed data structure, which contains the related analog parameters and digital control bits. Further, success flags indicate whether the calibration targets were reached within the available parameter ranges. These result structures can either directly be applied to a hardware setup or serialized to disk. Application of serialized calibration is beneficial compared to repeating the calibration in experiments due to decreased required time and improved digital reproducibility.

Running the calibration on a host computer using `Python` allows for great flexibility in terms of gathering observations from the chip. We can utilize all observables, including a fast ADC, which allows performing fits to measured data – as sketched previously for the calibration of the membrane time constant. While this direct measurement should yield the most accurate results, fitting to a trace for each neuron takes a lot of time. Performing a full LIF neuron calibration takes a few minutes via the `Python` module. And importantly, when scaling this approach to many chips, we need to scale the host computing power accordingly.

In order to achieve better scalability, we can control the calibration from the embedded processors, directly on chip, removing the host computer from the loop. However, this approach limits the observables to those easily accessible to the embedded processor, the CADC and spike counters – performing a fit to an MADC trace using the embedded processors would consume lots of runtime and potentially counteract benefits of scaling. As a result, some calibrations have to rely on an indirect measurement of their observable. Again using the neurons' membrane time constant as an example, we can consider the spike rate in a leak-over-threshold setup. However, this introduces a dependency on multiple potentials being calibrated beforehand.

Apart from the need for indirect measurements, on-chip and host-based calibration work similarly: An iterative algorithm selects parameters, we configure them on chip and characterize their effects. Using the embedded processors for configuring parameters and acquiring data from the two on-chip readouts is fully supported and naturally faster than fetching them from a host computer. We use the SIMD CPUs' vector units for parallel access to the synapse array and columnar ADCs. This is enabled by cross-compiler-support (cf. section 3.3), by which both the scalar unit and vector unit are integrated and accessible from the C++ language.

We provide routines for on-chip calibration, which allow all LIF neuron parameters to be calibrated in approximately half a minute, with this number staying constant even when considering large systems comprising many individual chips. Similar to the host-based calibration API, *calix* exposes these on-chip routines as conveniently parametrized functions that can be called within any experiment. Their runtime is mostly limited by waiting for configured analog parameters to stabilize before evaluating the effects on the circuits.

## 2.4 Platform Operation

Over the past decade neuromorphic systems evolved from intricate lab setups towards back ends for the more comfortable execution of spiking neural networks (Indiveri et al., 2011; Furber et al., 2012; Benjamin et al., 2014; Davies et al., 2018; Pehle et al., 2022). One major step along this development path is to provide users with seamless access to the systems.

Small scale prototype hardware is often connected to a single host machine, e.g., via USB. This is also a common usage mode for different neuromorphic hardware. To access these devices, users have to have (interactive) access to the particular machine the hardware is connected to. This limits the flexibility of the user and is an operational burden as the combination of neuromorphic hardware and host machine has to be maintained. While this tightly coupled mode of operation is sufficient during commissioning and initial experiments, it is not robust enough for higher work-loads and flexible usage.

An improvement to the situation sketched above is using a scheduler, e.g., SLURM (Yoo et al., 2003), where users can request a resource, e.g., a specific hardware setup, and the jobs get launched on the matching machine with locally attached hardware. This is the typical mode of access also used for other accelerator-type hardware, e.g., GPU clusters. However, this batch driven way is not always ideal as it often requires accounts on the local compute cluster and does not allow for easy interactive usage. In addition, traditional compute load schedulers optimize for throughput and not latency, therefore the scheduling overhead can be significant especially for hardware that is fast and experiments that are short. In the latter case, job execution rates of the order of Hz and faster are required.

Another downside of using a traditional scheduler is that hardware resources are not efficiently utilized when multiple users want to use the same hardware resources at the same time. Therefore, we developed the micro scheduler *quiggeldy* that exposes access to the hardware directly via a network connection, but still manages concurrent access from different users. It decouples the hardware utilization from the user's surrounding computations such as experiment preparation, updates in iterative workflows or result evaluation. For this to work runtime control, configuration, input stimulus as well as output data must be serializable which is facilitated via cereal (Grant and Voorhies, 2017). The inter-process communication between the user software and the micro scheduler is done with RCF (Delta V Software, 2020). When a user requests multiple hardware runs, it is checked whether certain already performed parts can be omitted, e.g., resets or re-initializations. Experiment interleaving between multiple users is also supported as the initialization state is tracker for each user and is automatically applied when needed.

Having the correct software environment for using neuromorphic hardware is also a major challenge. Nowadays, software vendors often provide a container image that includes the appropriate libraries. However, this approach does not necessarily yield well specified and traceable dependencies, but only a "working" black-box solution. We overcome this downside by using the Spack (Gamblin et al., 2015) package manager with a meta-package that explicitly tracks all software dependencies and their version needed to run experiments on and develop for the neuromorphic hardware. An automatically built container embedding the Spack installation enables encapsulation and eased distribution. This Spack meta-package is also used for the EBRAINS' JupyterLab service and will eventually be deployed to all HPC sites involved in EBRAINS (ebr, 2022). The latter will facilitate multi-site workflows involving neuromorphic hardware and traditional HPC.

# 3 Applications

In this section, we show-case a range of applications of BSS-2. Each application involves use of unique hardware features or modes of operation and motivates parts of the software design.

First, we describe biological multi-compartmental modelling in section 3.1 concluding in the development of an API for structural neurons. Continuing, functional modelling with spiking neural network (SNN) is demonstrated for a pattern-generation task in section 3.2, which leads to embedding of spiking BSS-2 usage into the machine learning framework PyTorch and involves host-based training as well as local learning on the SIMD CPUs. Then, embedded operation, where the SIMD CPUs are the experiment orchestrator of BSS-2, is displayed and their implications detailed in section 3.3. Following, the non-spiking mode of operation implementing ANNs and its PyTorch interface is characterized in section 3.4. Afterwards, user adoption and platform access to BSS-2 is shown in section 3.5. Finally, application of the software stack for hardware co-simulation, co-design and verification is portrayed in section 3.6.

## 3.1 Biological Modeling Example

BSS-2 aims to emulate biological inspired neuron models. Most neurons are not simple point-like structures but possess intricate dendritic structures. In recent years, the research interest in how dendrites shape the output of neurons has increased (Major et al., 2013; Gidon et al., 2020; Poirazi and Papoutsi, 2020). As a result, BSS-2 incorporates the possibility to emulate multi-compartmental neuron models in addition to the AdEx point-neuron model (Aamir et al., 2018; Kaiser et al., 2021).

In the following, we use a dendritic branch, which splits into two sub-branches, to illustrate how multi-compartmental neuron models are represented in our system, cf. fig. 9. At first, we look at a simplified representation of the model, subfigure (A). The main branch consists of two compartments, connected via a resistance; at the second compartment, the branch splits in two sub-branches, which themselves consist of two compartments each. On hardware this model is replicated by connecting several neuron circuits via switches and tunable resistors, cf. fig. 9 (B). Each compartment consists of at least two neuron circuits, directly connected via switches, compare colors in subfigure (A) and (B). With the help of a dedicated line at the top of the neuron circuits these compartments can then be connected via resistors to form the multi-compartmental neuron model; for more details see Kaiser et al. (2021).

In software, the `AtomicNeuron` class stores the configuration of a single neuron circuit and therefore can be used to configure the switches and resistors as desired. As mentioned in section 2.3.4, the `HXNeuron` exposes this data structure to the high-level interface PyNN, allowing users to construct multi-compartmental neuron models in a known environment. However, it is cumbersome and error-prone to set individual switches. As a consequence, we implement a dictionary-like hierarchy on top of the `AtomicNeuron`, called `LogicalNeuron` in the logical abstraction layer, cf. section 2.2.

We use a builder pattern approach to construct these logical neurons: the user creates a neuron morphology by defining which neuron circuits constitute a compartment and how these compartments are connected. Upon finalization of the builder, the correctness of the neuron model configuration of the neuron model is checked; if the provided configuration is valid, a `LogicalNeuron` is created. This `LogicalNeuron` stores the morphology of the neuron as well as the configuration of each compartment.

The coordinate system of the BSS-2 software stack, cf. section 2.2.2, allows to place the final logical neuron at different locations on the chip (Müller et al., 2020a). This is achieved by saving the relation between the different neuron circuits defining the morphology in relative coordinates. Once the neuron is placed at a specific location on the chip, the relative coordinates are translated to absolute coordinates.

Currently, the logical neuron is only exposed in the logical abstraction layer. In future work, it will be integrated in the PyNN API of the BSS-2 system. This will – for instance – allow to easily define populations of multi-compartmental neurons and connections between them.

### 3.2 Functional Modeling Example

The BSS-2 system enables energy efficient and fast SNN implementations. Moreover, the system's embedded SIMD CPU enables highly parallelized on-chip learning with fast access to observables and thus, promises to benefit the computational neuroscience and machine learning community in terms of speed and energy consumption. We demonstrate functional modeling on the BSS-2 system with a pattern-generation task using recurrent spiking neural networks (RSNNs) with an input layer, a recurrent layer and a single readout neuron. The recurrent layer consists of 70 LIF neurons $\{j\}$ with membrane potential $v_j^t$, receiving spike trains $x_i^t$ from 30 input neurons $\{i\}$. Neurons in the recurrent layer project spike events $z_j^t$ onto the single leaky-integrate readout neuron with potential $y^t$.

RSNNs are commonly trained using backpropagation through time (BPTT) by introducing a variety of surrogate gradients taking account of the discontinuity of spiking neurons (Bellec et al., 2020; Zenke and Ganguli, 2018; Shrestha and Orchard, 2018). However, as BPTT requires knowledge of all network states along the time sequence in order to compute weight updates (backwards locking), it is not just considered implausible from a biological perspective, but also unfavourable for on-chip learning, which effectively enables high scalability due to local learning. Therefore, we utilize e-prop learning rules (Bellec et al., 2020), where the gradient for BPTT is factorized into a temporal sum over products of so-called learning signals $L_j^t$ and synapse-local eligibility traces $e_{ji}^t$. While the latter accumulates all contributions to the gradient that can be computed forward in time, the first depends on the network's error and still requires BPTT. However, Bellec et al. (2020) provide suitable approximations for $L_j^t$, allowing computing the weight updates online (fig. 10A). Such learning rules are favorable for the BSS-2 system, as the SIMD CPU can compute the weight updates locally while the network is emulated in parallel.

E-prop-inspired learning on the BSS-2 system is enabled by adapting Bellec et al. (2020, Eq. (28)). Here we replace the membrane potentials $v_j^t$ in $e_{ji}^t$ with the post-synaptic recurrent spike train $z_j^t$,

$$e_{ji}^{t+1} \to z_j^{t+1} \cdot \mathcal{F}_\alpha \left( z_i^t \right) := \hat{e}_{ji}^{t+1}, \qquad \Delta W_{ji}^{\mathrm{hh}} = -\eta \sum_t L_j^t \mathcal{F}_\kappa \left( \hat{e}_{ji}^t \right), \qquad (1)$$

where $\mathcal{F}_x$ is an exponential filter with decay constant $x$. The update rule for input weights, derived in Bellec et al. (2020), is adapted accordingly. The equation for output weights remains untouched. With the readout neuron's membrane trace $y^t$ and a MSE loss measuring the error to a target trace $y^{*,t}$, the learning signals are $L_j^t = W_j^{\mathrm{ho}} \left( y^t - y^{*,t} \right)$. Since this learning rule propagates only spike-based information over time we refer to it as *s-prop*.

Finally, we approach s-prop learning with BSS-2 in the loop (cf. section 2.1). For this, the network, represented by PyTorch parameters $W^{\mathrm{ih,\ hh,\ ho}}$, is mapped to a hardware representation (see fig. 10B) via *hxtorch* (see section 2.3.5), forwarding a spike tensor on-chip. Inherently, *grenade* (see section 2.3.2) applies a routing algorithm, finds a graph-based experiment description and executes it on hardware for a given time intervall. The routing algorithm allocates two adjacent hardware synapses for one signed synapse weight in software, one excitatory and one inhibitory. Further, *grenade* records the MADC-sampled readout trace $y^t$ and the recurrent spike trains $\mathbf{z}^t$. Both observables are returned as PyTorch tensors for weight optimization on the host side. Experiment results are displayed in fig. 10C.

Implementing s-prop on-chip requires the SIMD CPU to know and process explicit spike-times. As this comes with a high computational cost, the correlation sensors are utilized to emulate approximations of the spike-based eligibility traces $\hat{e}_{ji}^{t}$ in analog circuits, thereby freeing computational resources on the SIMD CPU. The correlation sensors model the eligibility traces under nearest-neighbor approximation (Friedmann et al., 2017) and are accessed by the SIMD CPU as an entity $c_{ji}^{n}$, accumulated over a period $P$. Hence, the time sequence is split into $N$ chunks of size $P$ and weight updates on the SIMD CPU are performed at times $t^n = nP + \tilde{t}$, with $n \in \mathbb{N}_0^{<N}$ (cf. fig. 10E) and $\tilde{t} \in [0, P)$ a random offset,

$$\Delta \bar{W}_{ij}^{\text{ih/hh}} = -\eta \sum_n L_j^n \mathcal{F}_{\hat{\kappa}}\left(c_{ji}^{\text{ih/hh},n}\right) \quad \text{and} \quad \Delta \bar{W}_{kj}^{\text{ho}} = -\eta \sum_n \left(y_k^n - y_k^{*,n}\right) \mathcal{F}_{\hat{\kappa}}\left(\zeta_j^n\right), \qquad (2)$$

with $\hat{\kappa} = \exp\left(-P/\tau_\text{m}\right)$ and $\zeta_j^n$ being the recurrent spike count in interval $n$. Due to the updates rules' accumulative nature, we refer to them as neuromorphic accumulative spike propagation (NASProp). Simulations in fig. 10D verify that NASProp endows RSNNs with the ability to solve the pattern-generation task reasonable well.

NASProp's SIMD CPU implementation effectively demonstrates full on-chip learning on the BSS-2 system. In high-level software, on-chip learning is implemented in a PyTorch model, defined in *hxtorch*, holding parameters for the network's projections. Its `forward` method implicitly executes the experiment on the BSS-2 system for a batch of input sequences. Currently, this model learning on-chip serves as a mere black box for the specific network at hand with a static number of layers, as for on-chip spiking networks the network's topology needs to be known upon execution. Therefore, this approach is considered a first step from common PyTorch models to spiking on-chip models.

As for in-the-loop learning, on forwarding a batch of inputs sequences, *grenade* maps the software network to a hardware representation with signed synapses and configures the chip accordingly. Moreover, before executing a batch element, *grenade* starts the plasticity kernel on the SIMD CPU, computing weight updates in parallel to the network's emulation. The plasticity rule implementation relies on *libnux* (cf. section 2.2.3 and utilizes the VU extension for accessing hardware observables (e.g. $c_{ji}^n$ and $y^n$) and computing weight updates row-wise in parallel, thereby fully exploiting the system's speed up factor.

In *hxtorch*, learning parameters are configured in a configuration object exposed to `Python`, which is injected to *grenade* and passed to the SIMD CPU before batch execution on hardware begins. As different projections in the network have different update rules, relying on population-specific observables, the network's representation on hardware (cf. fig. 10B) is communicated to the SIMD CPU. This allows for identifying signed hardware synapses and neurons with projections and populations on the SIMD CPU. Finally, before each batch element is executed, *grenade* has the ability to write trial-specific details onto the SIMD CPU (e.g. random offset $\tilde{t}$ and the synapse row to perform updates for). Hence, smooth on-chip learning is granted by reliable communication between little-endian host engine and the embedded big-endian SIMD CPU. For serialization of information from and to the SIMD CPU we deploy the C++ header-only library *bitsery* (Vinkelis, 2020), allowing for seamless transmission of objects between systems of differing endianness.

Due to changing hardware weights during on-chip training, the adjusted weights are reverse mapped to the software representation and stored in the network's parameter tensors. Therewith we utilize PyTorch's native functionality to load and store network parameters. Reverse network mapping is implemented in the *hxtorch* on-chip-learning model by accessing the hardware routing result and is performed implicitly in the model's `forward` method after experiment execution.

Successful implementations of plasticity rules for on-chip learning are facilitated by providing transparency of SIMD CPU programs by means for tracing and recording data. To that end, *libnux* (cf. section 2.2.3) facilitates logging of any information into a dedicated SIMD CPU memory region, easily accessed from the host engine. Moreover, logging can be redirected to the FPGA-controlled dynamic random-access memory (DRAM), effectively allowing extensive logging of whole learning processes and hardware observables.

### 3.3 Embedded Operation

Apart from operating BSS-2 tightly coupled to a host computer, the integrated microprocessors can act as system controllers. They can orchestrate the control flow of the experiment and undertake tasks within it. These tasks may include calibration routines, virtual environment simulation or optimizer loops. Embedding them in proximity to the neural network core yields latency and data-locality advantages. In the following, we describe three exemplary experiments that make exhaustive use of the embedded processors as system controllers.

First, Wunderlich et al. (2019) introduce an embedded environment simulation of a simplified version of the Pong video game on the SIMD CPU, see left panel in fig. 11. One of the two involved agents plays optimally by design, the other one is represented by a SNN on BSS-2. During the experiment, the latter is trained on-chip using a reward-based spike timing dependent plasticity (STDP) rule. This set-up therefore unites the control flow, virtual environment simulation and learning rule within a single program running on the integrated processors.

Second, Stradmann et al. (2021) describe the application of the BSS-2 system for inference of ANNs that detect atrial fibrillation in medical electrocardiogram (ECG) data. Targeting applications in energy efficient devices, they aim for as little periphery as possible and therefore let the embedded processors orchestrate all classification routines. The resulting tight loop between the analog inference engine and digital data in- and outputs allows for low classification latencies and high throughput of more than 3600 ECG traces per second.

Third, Schreiber et al. (2022) presents the emulation of an insect model with strong biological inspiration on BSS-2. The simplified brain model is embedded into an agent that is fed with stimuli from a simulated environment, see right panel in fig. 11. While the neural network is emulated as a SNN within the analog core, the agent itself as well as its virtual environment are both simulated on the SIMD CPU. The authors specifically challenge the virtual insects with a simple path integration task: As depicted in the right panel of fig. 11, a simulated swarm-out phase is followed by a period of free flight, where the agent is supposed to return to its nest. The complexity of this task and the comparably low number of involved neurons requires precisely controlled dynamics, which they achieve by integrating experiment specific on-chip calibration routines directly on the SIMD CPUs (cf. section 2.3.6).

Supporting these complex experiments on the embedded processors and their interaction with the controlling host computer poses specific requirements to the BSS-2 OS. Especially, a cross-compilation toolchain for the SIMD CPU is required.

As described in section 2.2.3, we therefore provide a cross-compiler based on `gcc` (GNU Project, 2018), which in addition to the processor's scalar unit also integrates its custom vector unit in `C++` (Müller et al., 2020a). Additional hardware specific functionality is encapsulated in the support library *libnux*. It abstracts access to configuration data and observables in the analog neural network core, like synaptic weights or correlation measurements. The exchange of such data with the host is facilitated by integration of the lean, cross-platform binary serialization library *bitsery* (Vinkelis, 2020).

For execution, the compiled programs need to be placed in system memory — in case of BSS-2, each SIMD CPU has direct access to 16 kB SRAM. For a complete calibration routine or complex locally simulated environments, this may not suffice. We therefore utilize the controlling FPGA as memory controller: It allows the on-chip processors to access externally connected DRAM with significantly larger capacity at the cost of higher latency. Programs for the embedded processor can place instructions and data onto both the internal SRAM and the external memory via compiler attributes. This allows fine-grained decisions about the access-latency requirements of specific instruction and data sections.

Similar to experiments designed for operation from the host system, embedded experiments often require reconfiguration of parts of BSS-2. The hardware abstraction layer introduced in the BSS-2 OS (cf. section 2.2.2) has therefore been prepared for cross-compilation on the embedded processors. As a result, the described container and coordinate system can be used in experiment programs running on the on-chip SIMD CPUs.

### 3.4 Artificial Neural Networks

The BSS-2 hardware supports a non-spiking operation mode which supports artificial neural networks (ANNs) implementing multiply–accumulate (MAC) operations (Weis et al., 2020). The operation within the analog core is sketched in fig. 12A. Each entry in the vector operand stimulates one or two rows of synapses, when using unsigned or signed weights, respectively. The activations have an input resolution of 5 bit, controlling the duration of synapses' activation. Similar to the spiking operation, synapses emit a current pulse onto the neurons' membranes depending on their weight, which has a resolution of 6 bit. We implement signed weights by combining an excitatory and an inhibitory synapse into one logical synapse. Once all entries in the input vector have been sent to the synapses, the membrane potential resembles the result of the MAC operations. It is digitized for all neurons in parallel using the CADC, yielding an 8 bit result resolution.

As a user interface, we have developed an extension to the PyTorch machine learning framework (Paszke et al., 2019), *hxtorch* (Spilger et al., 2020). It partitions ANN models into chip-sized MAC operations that are executed on hardware using *grenade*, see section 2.2.5. Apart from a special MAC program used for each multiplication, the majority of code is shared between spiking and non-spiking operation. With the leak term disabled, the neurons' membranes represent the integrated synaptic currents, as shown in fig. 12B. As the MAC operation lacks any real-time requirements, it is executed as fast as possible to optimize energy efficiency. In terms of circuit parameterization, this means we choose a small synaptic time constant in order for the membrane potential to stabilize quickly. Therefore, a subset of the existing spiking calibration routines can be reused here, cf. section 2.3.6. There is only one additional circuit – the encoding of input activations to activation times in synapse drivers – that needs to be calibrated.

Defining an ANN model in *hxtorch* works similar to PyTorch: The *hxtorch* module provides linear and convolutional layer classes as a replacement for their PyTorch equivalents. We introduce a few additional parameters controlling the specifics of hardware execution, e.g. the time interval between sending successive entries in the input vector to the synapse matrix, or the option to repeat the vector for efficacy scaling. This enables the user to optimize saturation effects when driving the input currents as well as the gain of the MAC operation for the particular experiment. For both we provide default values as a starting point. The activation function `ConvertingReLU` additionally converts signed 8 bit output activations into unsigned 5 bit input activations for the following layer by a bitwise right shift.

Trained deep neural network models can be transferred to BSS-2 by first quantizing them with PyTorch and subsequently mapping their weights to the hardware domain. For quantization, we need to consider the intrinsic gain factor of the hardware MAC operation.

Figure 13 shows an example application of a deep neural network with BSS-2, using the yin-yang dataset from Kriener et al. (2021). One of the three classes – yin, yang, or dot – are to be determined from four input coordinates $(x, y, 1-x, 1-y)$. The network is first trained with 32 bit floating point accuracy using PyTorch, achieving 98.9 % accuracy. After quantizing with PyTorch to the hardware resolution of 5 bit activations and 6 bit plus sign weights, this drops to 94.0 %. Porting the model to BSS-2, after running a few epochs of hardware-in-the-loop training, an accuracy of 95.8 % is finally reached.

In addition to running the ANN on the BSS-2 hardware, a hardware-limitations-aware simulation is available. It can be enabled per layer via the `mock` parameter (see fig. 13B). For mock mode, we simply assume a linear MAC operation, using a hardware-like gain factor. To investigate possible effects of the analog properties of the BSS-2 hardware on the inference and training, additional Gaussian noise of the accumulators and multiplicative fixed-pattern deviations in the weight matrix can be simulated. The comparison with actual hardware operation shown in fig. 13 D illustrates how this simple model already captures the most dominant non-linearities of the system. More sophisticated software representations that embrace second-order effects across multiple hardware instances have been proposed by Klein et al. (2021). They have shown how pre-training with faithful software models can significantly decrease hardware allocation time while at the same time increasing classification accuracy compared to plain hardware-in-the-loop training.

## 3.5 User Adoption and Platform Access

The BSS-2 software stack aims to enable researchers to exploit the capabilities of the novel neuromorphic substrate. Support for common modeling interfaces like PyNN and PyTorch provides a familiar entry point for a wide range of users. However, not all aspects of the hardware can fully be abstracted away, requiring users to familiarize themselves with unique facets of the system. To flatten the learning curve several tutorials —verified in continuous integration (CI) as 'executable' documentation— as well as example experiments are provided[17]. They range from introducing the hardware via single neuron dynamics to learning schemes like plasticity rate coding. In addition to the scientific community, they also target students, for example exercises accompanying a lecture about Brain Inspired Computing and hands-on tutorials.

A convenient entry point to explore novel hardware are interactive web-based user interfaces. That is why we integrated the BSS-2 system into the EBRAINS Collaboratory[18] (ebr, 2022). The Collaboratory provides a dynamic VM hosting on multiple HPC sites for Jupyter notebooks running in a comprehensive software environment. An BSS-2-specific experiment service manages multi-user access to the hardware located in Heidelberg utilizing the *quiggeldy* micro scheduler, see section 2.4. It allows for seamless interactive execution of experiments running on hardware with execution rates of over 10 Hz. This was, for example, utilized during hands-on tutorials at the NICE 2021 conference (nic, 2021) of which the execution rates are shown in fig. 14.

Furthermore, EBRAINS has begun to provide a comprehensive software distribution that includes typical neuroscientific software libraries next to the BSS-2 client software. As of now, this software distribution has been already deployed at two HPC centers and work is under way to extend this to all sites available in the EBRAINS community. Leaving interactive demos aside, this automatic software deployment will simplify multi-site workflows significantly —including BSS-2 systems— as the scientist is not responsible for software deployment anymore.

---

[17]The tutorials and example experiments are available at https://github.com/electronicvisions/brainscales2-demos

[18]Platform access is available via https://ebrains.eu

### 3.6 Hardware/Software Co-Development

The BSS-2 platform consists of two main hardware components: the ASIC implementing an analog neural network core and digital periphery, as well as a FPGA used for experiment control and digital communication. Development of these hardware components is primarily driven by simulations of their analog and digital behavior, where — especially in the case of the ASIC — solid pre-fabrication test strategies need to be employed. Given the complexity of the system, integration tests involving all subsystems are required to ensure correct behavior.

Replicating the actual hardware systems, the setup for these simulated integration tests pose very similar requirements on the configuration and control software. The BSS-2 OS therefore provides a unified interface to both, circuit simulators and hardware systems. For the connection to the simulators, we introduce an adapter library (*flange*) as an optional substitution for the network transport layer. Implementing an additional *hxcomm* back-end, *flange* allows for the transparent execution of hardware experiments in simulation.

This architecture enables various synergies between hardware and software development efforts — specifically, co-design of both components already in early design phases. On system level, this methodology helps to preempt interface mismatch between components of various different subsystems. Positive implications for software developers include the possibility of very early design involvement as well as enhanced debug information throughout the full product life cycle: Having simulation models of the hardware components of the system allows for the inspection of internal signals within the FPGA and ASIC during program runtime. In particular, we have made use of this possibility during the development of a compiler toolchain for the embedded custom SIMD microprocessors, where the recording of internal state helps to understand the system's behavior. Hardware development, on the other hand, strongly profits from software-driven verification strategies and test frameworks. BSS-2 OS especially allows to run the very same test suites on current hardware as well as simulations of future revisions. These shared test suites are re-used across all stages of the platform's life cycle for multiple hardware generations, therefore ever accumulating verification coverage.

## 4 Discussion

This work describes the software environment for the latest BrainScaleS (BSS) neuromorphic architecture (Pehle et al., 2022): the BrainScaleS-2 (BSS-2) operating system. In Müller et al. (2020b) we introduced the operating system for the BrainScaleS-1 (BSS-1) wafer-scale neuromorphic hardware platform. New basic concepts of the second-generation software architecture were described in Müller et al. (2020a). For example, we introduced a concise representation of "units of configuration" and "experiment runs" supporting asynchronous execution by extensive usage of future variables. Key concepts already existing in BSS-1 —e.g., the type-safe coordinate system— were extended for BSS-2. In particular, the systematic use of futures now allows higher software levels to transparently support experiment pipelining and asynchronous experiment execution in general. Additionally, dividing experiments into a definition and an execution phase also facilitates experiment correctness, software stack flexibility —by decoupling hardware usage from experiment definition— as well as increased platform performance by enabling a separation of hardware access from other aspects of the experiment.

The new software framework is expert-friendly: we designed the software layers to facilitate composition between higher- and lower-level application programming interfaces (APIs). Domain experts can therefore define experiments on a higher abstraction level in certain aspects, and are still able to access low-level functionality. A software package for calibration routines —the process of tuning hardware parameters to the requirements defined by an experiment— provides algorithms and settings for typical parameterizations of the chip, including support for multi-compartmental neurons and non-spiking use cases. An experiment micro scheduler service allows to pipeline experiment runs, and even preempt longer experiment sessions of individual users, to decrease hardware platform latency for other user sessions. Enabling multiple high-level modeling interfaces —such as PyNN and PyTorch— to cover a larger user base was one of the new requirements for BSS-2. To achieve this, we provide a separate high-level representation of user-defined experiments. This signal-graph-based representation is generally suited for high-level configuration validation, optimization, and transformation from higher- to lower-level abstractions. The modeling API wrappers merely provide conversions between data types and call semantics. The embedded microprocessors allow for many new applications: Initially designed to increase flexibility for online learning rules, (Friedmann et al., 2017), they have been also used for: environment simulations (Schreiber et al., 2022; Pehle et al., 2022), online calibration (section 3.3), general optimization tasks, as well as experiment control (Wunderlich et al., 2019). We ported our low-level chip configuration interface to the embedded processors and thereby allow for code sharing between host and embedded program parts in addition to a software library for embedded use cases. Apart from features directly concerning platform users, we enhanced the support for multiple hardware revisions in parallel facilitating hardware development, commissioning and platform operation. In combination with a dedicated communication layer, this enables not only support for multiple communication backends between host computer and field-programmable gate array (FPGA), such as gigabit ethernet (GbE) or a memory-mapped interface for hybrid FPGA-CPU systems, but also for co-simulation and therefore co-development of software and hardware. Finally, we operate BSS-2 as a research platform. As a result of our contributions to the design and implementation of the EBRAINS (ebr, 2022) software distribution, interactive usage of BSS-2 is now available to a world-wide research community. To summarize, we motivated key design decisions and demonstrated their implementation based on existing use cases: Support for multiple top-level APIs for 'biological' and 'functional' modeling; support for the embedded microprocessors including structured data exchange with the host, a multi-platform low-level hardware-abstraction layer, and an embedded execution runtime and helper library; support for artificial neural networks in host-based and standalone applications; focus on the user community by providing an integrated platform; sustainable hardware-software co-development.

To build a versatile modeling platform, BSS-2 is a neuromorphic system that improved upon successful properties of predecessors, both, in terms of hardware and software. Simulation speed continues to be an important point in computational neuroscience. The development of new approaches to numerical simulation promising lower execution times and better scalability is an active field of research (Knight and Nowotny, 2018, 2021; Abi Akar et al., 2019), as is improving existing simulation codes (Kunkel et al., 2014; Jordan et al., 2018). Whereas parameter sweeps scale trivially, systematically studying model dynamics over sufficiently long periods as well as iterative approaches to training and plasticity can only benefit from increases in simulation speed. The physical modeling approach of the accelerated neuromorphic architectures allows for a higher emulation speed than state-of-the-art numerical simulations (Zenke and Gerstner, 2014; van Albada et al., 2021). BSS-2 can serve as an accelerator for spiking neural networks and therefore opens up opportunities to work on scientific questions that aren't accessible by numerical simulation. However, to deliver on this promise in reality, both, hardware and software need to be carefully designed, implemented and applied. The publications building on BSS-2 are evidence of what is possible in terms of modeling on accelerated neuromorphic hardware (Bohnstingl et al., 2019; Billaudelle et al., 2020, 2021; Cramer et al., 2019, 2022; Czischek et al., 2022; Göltz et al., 2021b; Kaiser et al., 2021; Klassert et al., 2021; Klein et al., 2021; Müller et al., 2020a; Schreiber et al., 2022; Spilger et al., 2020; Stradmann et al., 2021; Weis et al., 2020; Wunderlich et al., 2019).

We believe that these offer a first glimpse of what will be possible in a scaled-up system. The next step on the roadmap is a multi-chip BSS-2 setup employing EXTOLL for host and inter-chip connectivity (Neuwirth et al., 2015; Resch et al., 2014). First multi-chip experiments have been performed on a lab setup (Thommes et al., 2022). Additionally, a multi-chip system reusing BSS-1 wafer-scale infrastructure is in the commissioning phase and will provide up to 46 BSS-2 chips. Similar to BSS-1, a true wafer-scale version of BSS-2 will provide an increase in terms of resources by one order of magnitude and thus will enable research that not only looks at dynamics at different temporal scales, but also on larger spatial scales. In terms of software we have been adapting our roadmap continuously to match modelers' expectations. For example, we work on future software abstractions that will allow for flexible descriptions of spiking network models with arbitrary topology in a machine learning framework. PyTorch libraries such as BindsNET (Hazan et al., 2018) or norse (Pehle and Pedersen, 2021) enable efficient machine-learning-inspired modeling with spiking neural networks and would benefit from neuromorphic hardware support.

# Acknowledgments

The authors wish to thank all present and former members of the Electronic Vision(s) research group contributing to the BrainScaleS-2 neuromorphic platform.

# Funding

This work has received funding from the EC Horizon 2020 Framework Programme under grant agreements 785907 (HBP SGA2) and 945539 (HBP SGA3), the Deutsche Forschungsgemeinschaft (DFG, German Research Foundation) under Germany's Excellence Strategy EXC 2181/1-390900948 (the Heidelberg STRUCTURES Excellence Cluster), the German Federal Ministry of Education and Research under grant number 16ES1127 as part of the *Pilotinnovationswettbewerb 'Energieeffizientes KI-System'*, the Helmholtz Association Initiative and Networking Fund [Advanced Computing Architectures (ACA)] under Project SO-092, as well as from the Manfred Stärk Foundation, and the Lautenschläger-Forschungspreis 2018 for Karlheinz Meier.

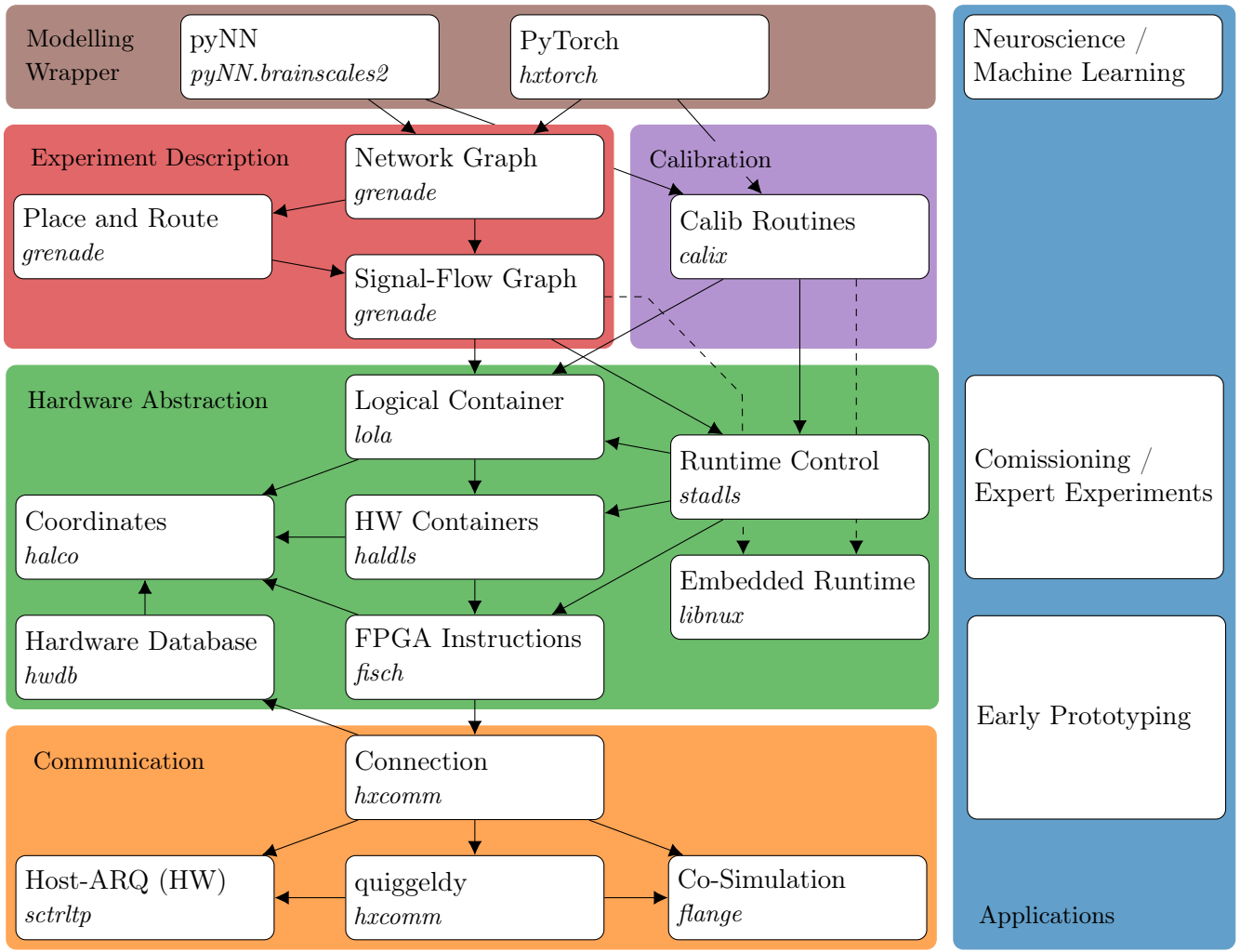

Figure 3: Overview of the BSS-2 software architecture and its applications. Left side: Colored boxes in the background represent the separation of the software into different concerns. White boxes represent individual software APIs or libraries with their specific repositories names and dependencies. Right side: Various applications concerning different system aspects. The arrows represent dependencies in the stack, where the dependent points to its dependencies. For embedded operation additional dependencies on *libnux* are needed (dashed arrows).

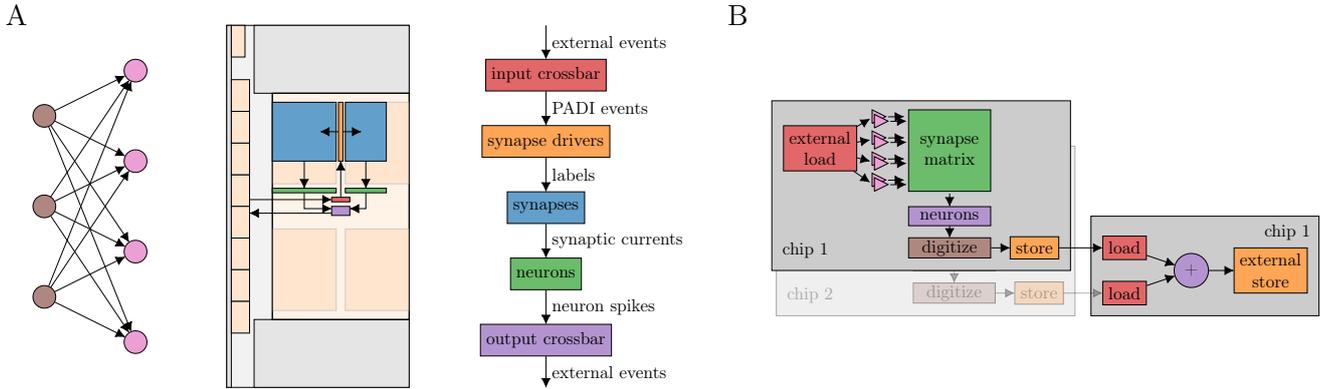

Figure 4: Signal-flow graph-based experiment abstraction on BSS-2. **(A)** Placed feed-forward network represented as signal-flow graph. Left: Abstract network; Middle: Actual layout on the chip, the arrows represent the graph edges; Right: The network graph structure enlarged with signal type annotation on the edges. **(B)** Non-spiking network distributed over two physical chips, adapted from Spilger et al. (2020). The result of two matrix multiplications on chips 1 and 2 is added on chip 1. The latter execution instance depends on the output of the two former instances.

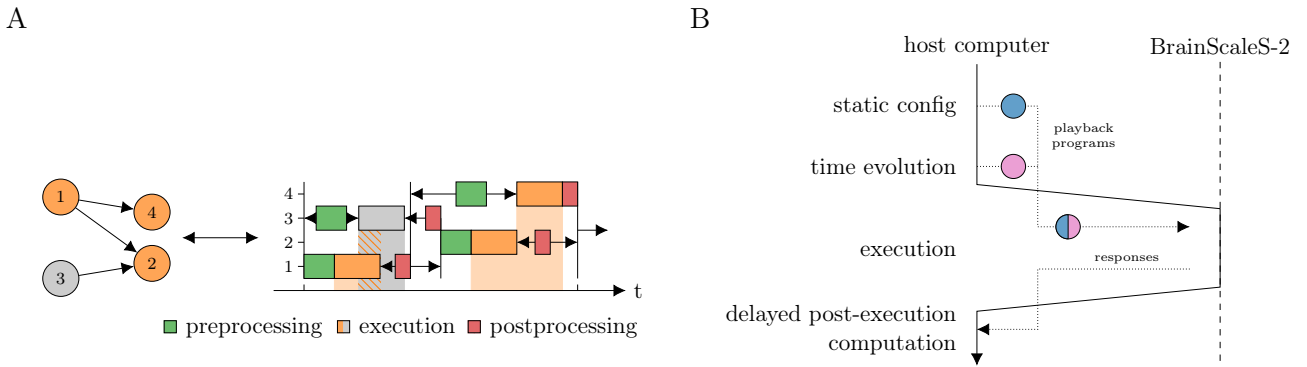

Figure 5: JIT compilation and execution of signal-flow graph of multiple execution instances and within a single execution instance. **(A)** JIT execution of a graph on two physical chips, adapted from Spilger et al. (2020). Left: Execution instance 3 is to be executed on another physical chip than the other execution instances. Right: The execution of instance 3, depicted in gray, can be performed concurrently to execution instance 1. **(B)** JIT compilation and execution of a single execution instance subgraph. First, the static configuration is extracted by a vertex visit and transformed to hardware configuration where applicable. Then, the real-time execution is built by a vertex visit. This built program is executed on the neuromorphic hardware and results are transmitted back to the host computer. Finally, delayed digital operations, which require output data from the execution, are performed on the host computer.

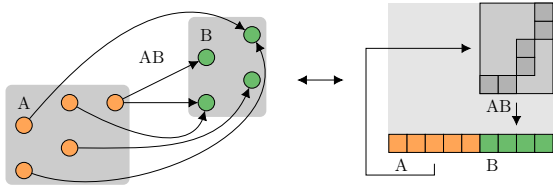

Figure 6: Abstract network notation. Population A consisting of five neurons is connected to population B consisting of four neuron via projection AB. Left: Abstract network; Right: Placed and routed on the hardware, where the projection AB consists of synapses in the two-dimensional synapse matrix and the populations A and B are located in the neuron row, compare fig. 1 (D).

```
neuron = lola.AtomicNeuron()
neuron.leak.v_leak = 650
neuron.leak.i_bias = 420
neuron.leak.enable_division = True
```

```
pynn.Population(1, pynn.HXNeuron({
  "leak_v_leak": 650,
  "leak_i_bias": 420,
  "leak_enable_division": True}))
```

Figure 7: Comparison between `lola.AtomicNeuron` and `pynn.HXNeuron`.

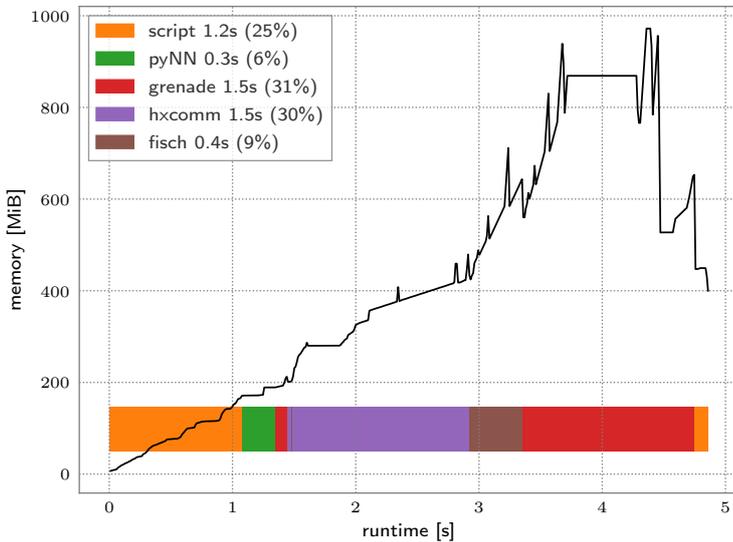

Figure 8: Run time analysis of a PyNN-based experiment with large spike count. Population of 12 neurons is excited by a regular spike train with frequency of $1\,\mathrm{MHz}$. The network is emulated for $1\,\mathrm{s}$ on hardware resulting in $1.2 \times 10^7$ spike events. The black line represents memory consumption during execution. Horizontal bars represent time consumption in software layers. The annotations in the legend present the individual run time of steps and percentage of the overall run time.

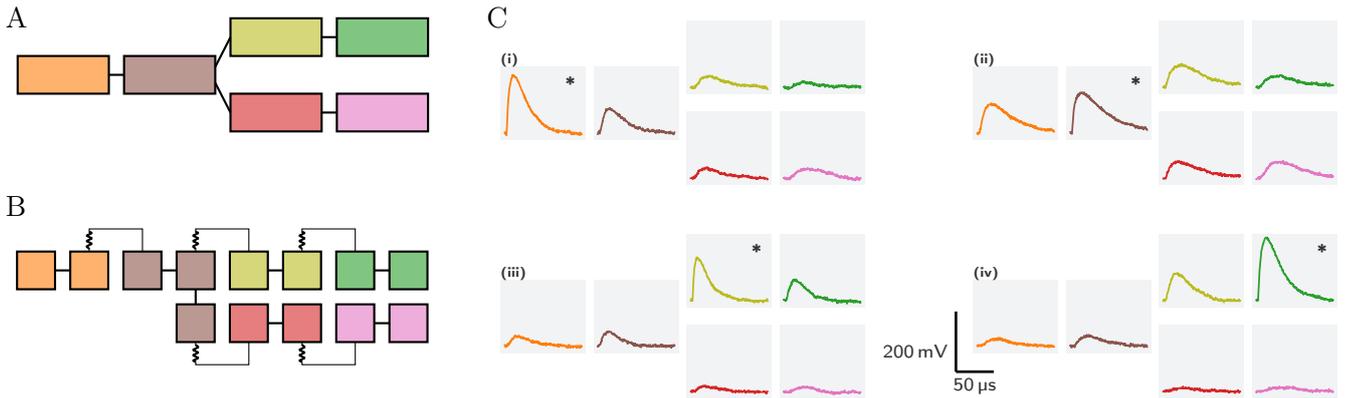

Figure 9: Pulse propagation along a dendrite which branches into two sub-branches. **(A)** Each branch is modeled by two compartments (rectangles). Different compartments are connected via resistors (lines). **(B)** Hardware configuration: neuron circuits (squares) are arranged in two rows on BSS-2, compare fig. 1 (D). Each compartment is represented by at least two neuron circuits. Circuits which form a single compartment are directly connected via switches (straight lines); compartments are connected via resistors. For details see Kaiser et al. (2021). **(C)** Membrane responses to synaptic input: we inject synaptic input at four different compartments; the compartment at which the input is injected is marked by a *. The membrane traces of the different compartments are arranged as in sub-figure (A). For the top left quadrant (i) the input is injected in the first compartment and decreases in amplitude while it travels along the chain. The response in both branches is symmetric. A similar behavior can be observed when the input is injected in the second compartment, (ii). Due to the symmetry of the model, we only display membrane responses for synaptic input to the upper branch. When injecting the input in the first compartment of the upper branch (iii) the input causes a noticeable depolarization within the same branch and the main branch but does not cause a strong response in the lower sister branch. Note: all values are given in the hardware domain.

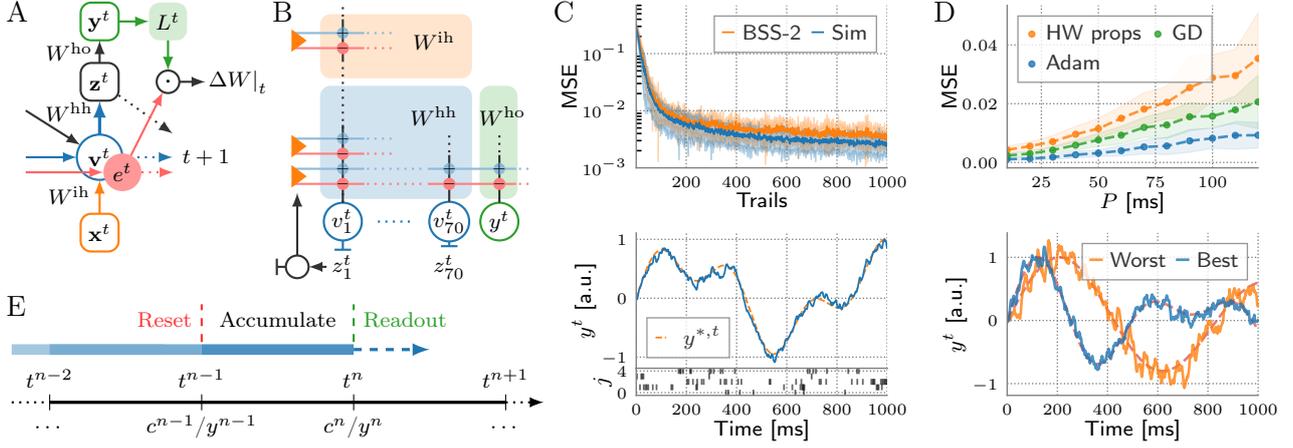

Figure 10: **(A)** Computational graph of an RSNN for one time step. The contribution to the weight update is computed by merging learning signals $L_j^t$ with eligibility traces $e_{ji}^t$. **(B)** Representation of the RSNN on the BSS-2 system using signed synapses. Inputs and recurrent spike trains are routed to the corresponding synapse drivers via the crossbar. **(C)** s-prop training on hardware. The upper plot depicts the evolution of the mean squared error (MSE) while training the BSS-2 system in-the-loop in comparison to training with the network simulated in software, incorporating basic hardware properties (Sim). In both cases the weights are optimized using the Adam optimizer (Kingma and Ba, 2014). The learned analog membrane trace of the readout neuron after training BSS-2 for 1000 epochs is exemplified in the lower plot, aligned to the spike trains $z_j^t$ of the first five out of 70 recurrent neurons. **(D)** NASProp simulations. The upper plot depicts the MSE over the update period $P$ after training with Adam in comparison to a training with gradient descent (GD) and a training taking additional hardware properties (noise, weight saturation, etc.) into account (HW props). Optimization with pure GD mimics weight updates computed by the SIMD CPU while on-chip learning. The lower plot shows the worst and best learned readout traces of the target pattern ensemble in simulation. **(E)** Timing of NASProp weight updates. For each update $n$ at $t^n$, the correlation $c_{ji}^n$ are merged with the learning signals $L_j^n$ by incorporating the membrane trace $y^n$.

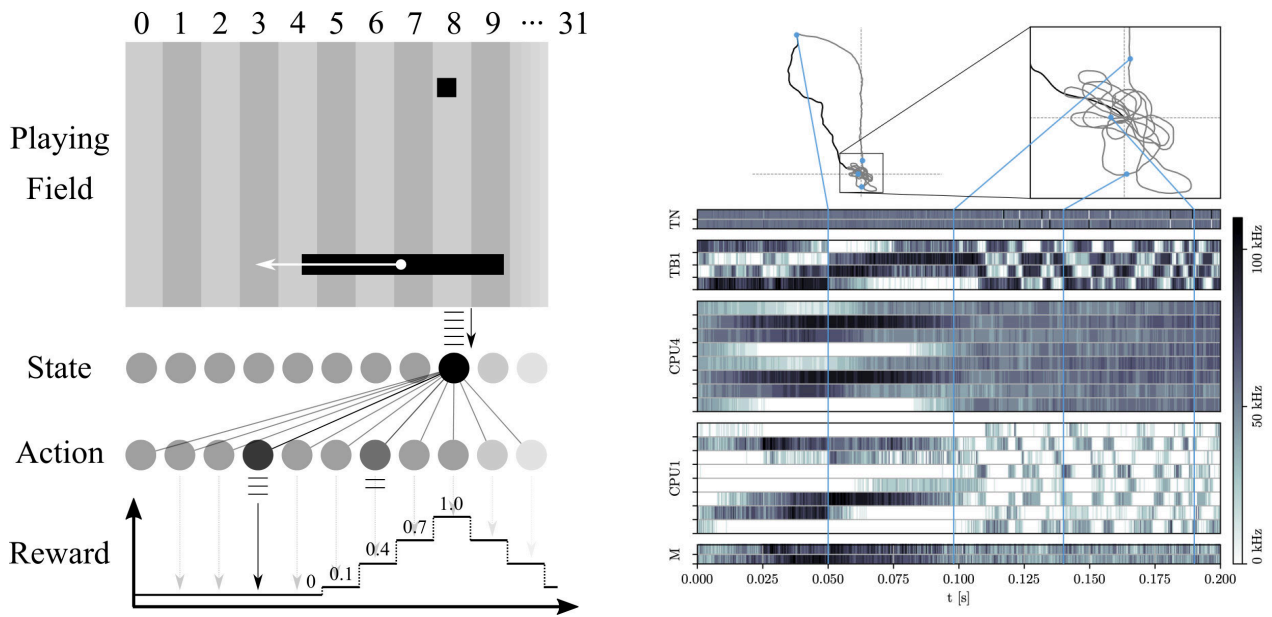

Figure 11: **(left)** Reinforcement learning: the chip implements a spiking neural network sensing the current ball position and controlling the game paddle. It is trained via a reward-based STDP learning rule to achieve almost optimal performance. The game environment, the motor command and stimulus handling, the reward calculation and the plasticity is performed by a `C++` program running on the on-chip processor. Figure taken from Wunderlich et al. (2019). **(right)** Recording of a virtual insect navigating a simulated environment. The top panels show the forced swarm-out path in black. During this phase, the SNN emulated by the analog neuron and synapse circuits on BSS-2 perform path integration. Afterwards, the insect flies freely and successfully finds its way back to the starting point and circles around it (gray trajectory). The bottom panel shows the neuronal activity during the experiment. The environment simulation as well as the interaction with the insect is performed by a `C++` program running on the on-chip processor. Figure taken from Pehle et al. (2022).

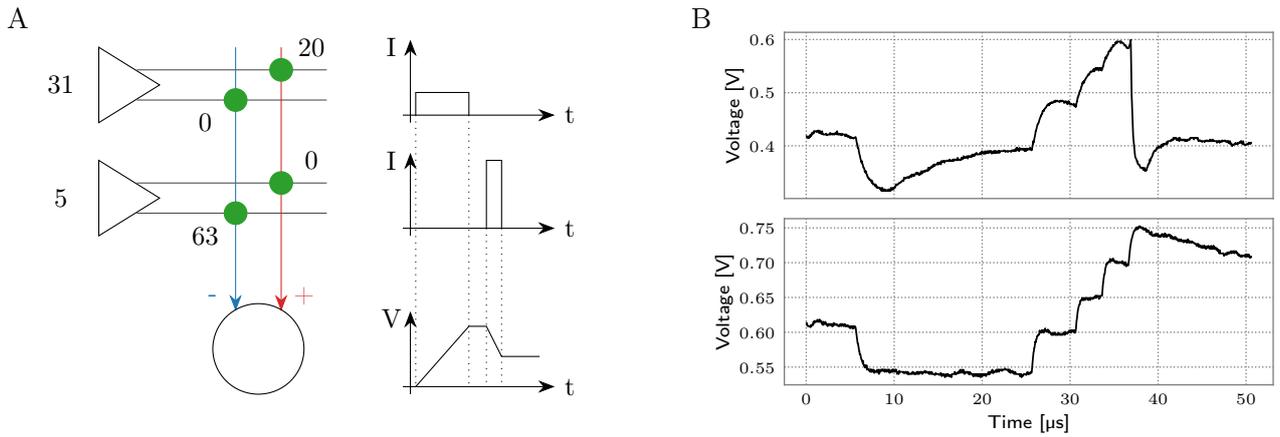

Figure 12: Matrix-vector multiplication for ANN inference. **(A)** Scheme of a multiply-accumulate operation. Vector entries are input via synapse drivers (left) in 5 bit resolution. They are multiplied by the weight of an excitatory or inhibitory synapse, yielding 6 bit plus sign weight resolution. The charge is accumulated on neurons (bottom). Figure taken from Weis et al. (2020). **(B)** Comparison between a spiking (top) and an integrator (bottom) neuron. Both neurons receive identical stimuli, one inhibitory and multiple excitatory inputs. While the top neuron shows a synaptic time constant and a membrane time constant, the lower is configured close to a pure integrator. We use this configuration for ANN inference. Please note that for visualization purposes the input timing (bottom) has been slowed to match the SNN configuration (top). The integration phase typically lasts less than 2 µs.

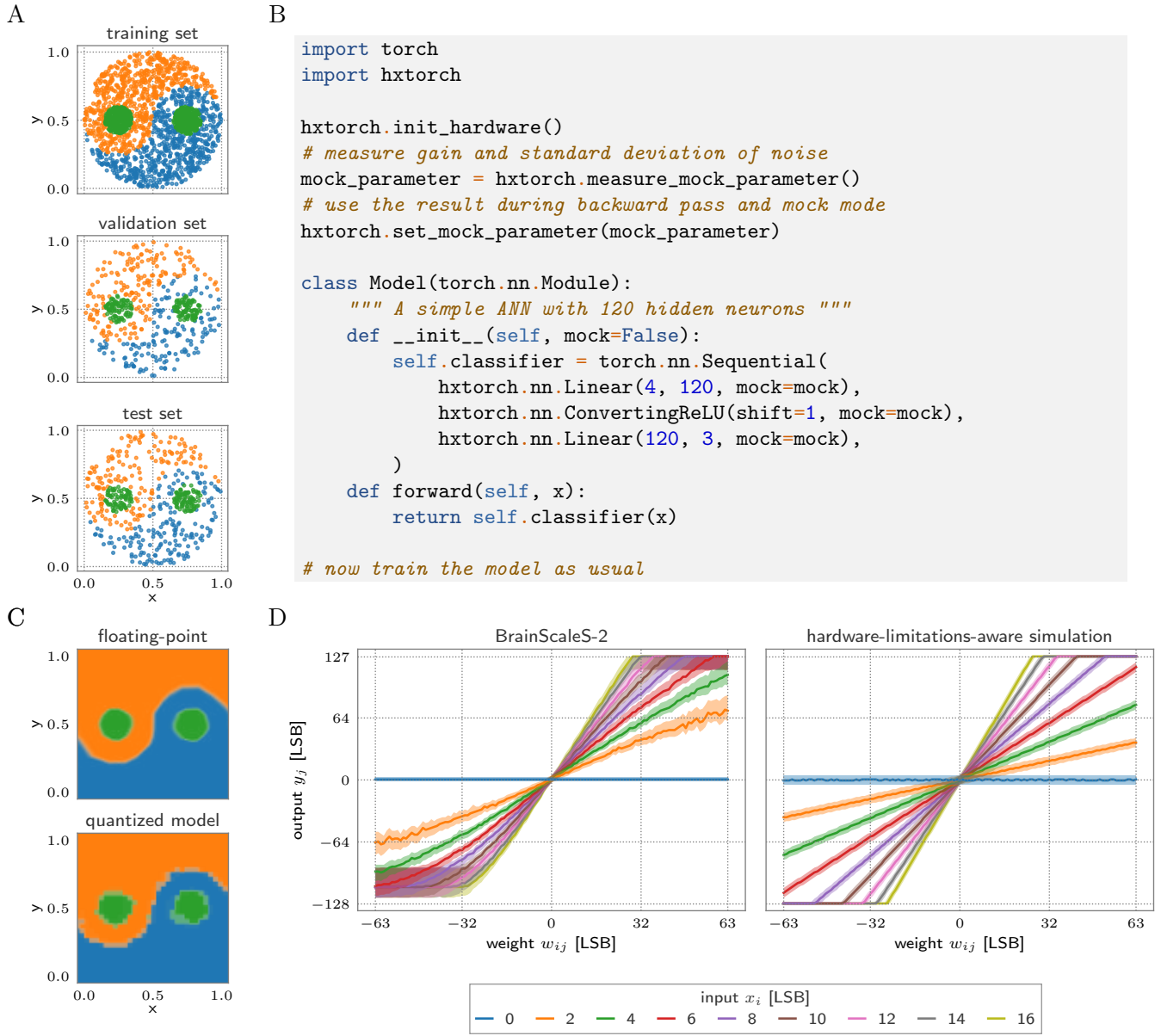

Figure 13: **(A)** The Yin-Yang dataset (Kriener et al., 2021) used for the experiment. **(B)** Hardware initialization and model description with *hxtorch*. **(C)** Network response of the trained model depending on the input. Top: 32 bit floating-point precision; bottom: quantized model on BSS-2 (5 bit activations, 6 bit plus sign weights). **(D)** Output of the MAC operation on BSS-2 (left) compared to the linear approximation (right). The solid line indicates the median, the colored bands contain 95% of each neuron's outputs across 100 identical MAC executions.

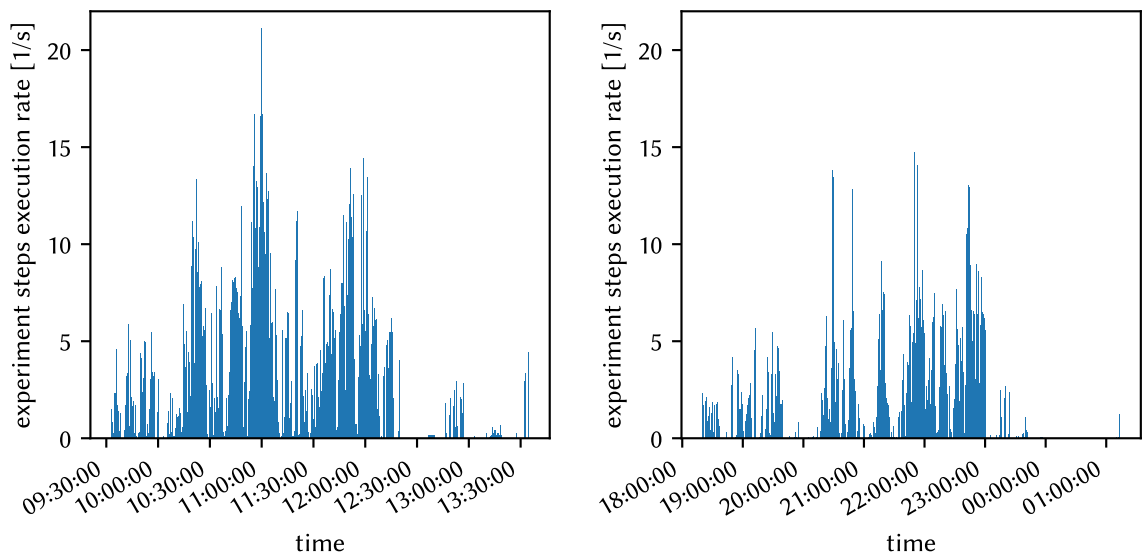

Figure 14: Rate of executed experiment-steps via *quiggeldy* during the two BSS-2 hands-on tutorials at NICE 2021. Experiments were distributed among eight hardware setups. In total there were 86 077 hardware runs executed.

43